\documentclass{article}
\usepackage{spconf,amsmath,graphicx}
\usepackage{times}
\usepackage{helvet}
\usepackage{courier}
\usepackage{xcolor}
\usepackage{graphicx}
\usepackage{multicol}
\usepackage{multirow}
\usepackage{pifont}
\usepackage{dsfont}
\usepackage{hyperref}
\usepackage{amsmath, amsfonts, amssymb, amsthm, bm}
\usepackage{algorithm}
\usepackage[noend]{algpseudocode}
\usepackage{multirow}

\newtheorem{theorem}{Theorem}

\theoremstyle{definition}

\theoremstyle{remark}

\DeclareMathOperator*{\expct}{\mathbb{E}}
\DeclareMathOperator*{\argmax}{arg\,max}
\DeclareMathOperator*{\argmin}{arg\,min}

\def \S {\mathcal{S}}
\def \H {\mathcal{H}}
\def \I {\mathbb{I}}
\def \X {\mathcal{X}}
\def \Y {\mathcal{Y}}

\def \eps {\epsilon}
\def \del {\delta}

\newcommand{\bra}[1]{\left\{ #1 \right\}}
\newcommand{\sbra}[1]{\left[ #1 \right]}
\newcommand{\rbra}[1]{\left( #1 \right)}
\renewcommand{\vec}[1]{\mathbf{#1}}
\newcommand{\overbar}[1]{\mkern 1.5mu\overline{\mkern-1.5mu#1\mkern-1.5mu}\mkern 1.5mu}
\newcommand{\rpm}{\sbox0{$1$}\sbox2{$\scriptstyle\pm$}
  \raise\dimexpr(\ht0-\ht2)/2\relax\box2 }

\newcommand*\samethanks[1][\value{footnote}]{\footnotemark[#1]}

\title{ON THE POWER OF DEEP BUT NAIVE PARTIAL LABEL LEARNING}
\name{Junghoon Seo$^{1 \dagger}\thanks{$\dagger$ Both authors contributed equally to this work.}$\qquad\qquad Joon Suk Huh$^{2  \dagger}$\samethanks}
\address{
  $^{1}$SI Analytics Co. Ltd, South Korea \quad\quad
  $^{2}$UW--Madison, USA \\
{\small \tt jhseo@si-analytics.ai \quad \qquad\qquad \small \tt jhuh23@wisc.edu}
\\}
\begin{document}
%
\maketitle
\begin{abstract}
Partial label learning (PLL) is a class of weakly supervised learning where each training instance consists of a data and a set of candidate labels containing a unique ground truth label.
To tackle this problem, a majority of current state-of-the-art methods employs either label disambiguation or averaging strategies.
So far, PLL methods without such techniques have been considered impractical. In this paper, we challenge this view by revealing the hidden power of the oldest and naivest PLL method when it is instantiated with deep neural networks.
Specifically, we show that, with deep neural networks, the naive model can achieve competitive performances against the other state-of-the-art methods, suggesting it as a strong baseline for PLL.
We also address the question of how and why such a naive model works well with deep neural networks.
Our empirical results indicate that deep neural networks trained on partially labeled examples generalize very well even in the over-parametrized regime and without label disambiguations or regularizations.
We point out that existing learning theories on PLL are vacuous in the over-parametrized regime. Hence they cannot explain why the deep naive method works.
We propose an alternative theory on how deep learning generalize in PLL problems. 
\end{abstract}
\begin{keywords}
classification, partial label learning, weakly supervised learning, deep neural network, empirical risk minimization
\end{keywords}
\section{Introduction}
State-of-the-art performance of the standard classification task is one of the fastest-growing in the field of machine learning. In the standard classification setting, a learner requires an unambiguously labeled dataset. However, it is often hard or even not possible to obtain completely labeled datasets in the real world. 
Many pieces of research formulated problem settings under which classifiers are trainable with incompletely labeled datasets. These settings are often denoted as \textit{weakly supervised}.  \textit{Learning from similar vs. dissimilar pairs} \cite{hsu2019multi}, \textit{Learning from positive vs. unlabeled data} \cite{kiryo2017positive,kaji2018multi}, \textit{Multiple instance learning} \cite{maron1998framework,wang2019comparison} are some examples of weakly supervised learning.

In this paper, we focus on \textit{Partial label learning} \cite{jin2003learning} (PLL), which is one of the most classic examples of weakly supervised learning. In the PLL problem, classifiers are trained with a set of candidate labels, among which only one label is the ground truth. Web mining \cite{luo2010learning}, ecoinformatic \cite{liu2012conditional}, and automatic image annotation \cite{zeng2013learning} are notable examples of real-world instantizations of the PLL problem.

The majority of state-of-the-art parametric methods for PLL involves two types of parameters.
One is associated with the label confidence, and the other is the model parameters. These methods iteratively and alternatively update these two types of parameters. This type of methods is denoted as \textit{identification-based}.
On the other hand, \textit{average-based} methods \cite{hullermeier2006learning,cour2011learning} treat all the candidate labels equally, assuming they contribute equally to the trained classifier.
Average-based methods do not require any label disambiguation processes so they are much simpler than identification-based methods.
However, numerous works \cite{jin2003learning,zhang2015solving,tang2017confidence,feng2019partiala,DBLP:conf/aaai/YanG20} pointed out that the label disambiguation processes are essential to achieve high-performance in PLL problems, hence, attempts to build a high-performance PLL model through the average-based scheme have been avoided.

Contrary to this common belief, we show that one of naivest and oldest average-based methods can train accurate classifiers in real PLL problems. Specifically, our main contributions are two-fold:
\begin{enumerate}
  \item We generalize the classic naive model of \cite{jin2003learning} to the modern deep learning setting. Specifically, we present a naive surrogate loss for deep PLL. We test our deep naive model's performance and show that it outperforms the existing state-of-the-art methods despite its simplicity.\footnote{All codes for the experiments in this paper are public on \url{https://github.com/mikigom/DNPL-PyTorch}.}
  \item We empirically analyze the unreasonable effectiveness of the naive loss with deep neural networks. Our experiments shows closing generalization gaps in the over-parametrized regime where bounds from existing learning theories are vacuous. We propose an alternative explanation of the working of deep PLL based on observations of Valle-Perez et al. \cite{valle2018deep}.
\end{enumerate}

\section{Deep Naive Model for PLL}

\subsection{Problem Formulation}
We denote $x\in\X$ as a data and $y\in\Y=\bra{1,\dots,K}$ as a label, and a set $S\in\S=2^{\Y}\setminus\emptyset$ such that $y\in S$ as a partial label.
A partial label data distribution is defined by a joint data-label distribution $p(x,y)$ and a partial label generating process $p(S|x,y)$ where $p(S|x,y)=0$ if $y\notin S$.
A learner's task is to output a model $\theta$ with small $\text{Err}(\theta)=\expct_{(x,y)\sim p(x,y)}\I\rbra{h_\theta(x)\neq y}$ given with a finite number of partially labeled samples $\bra{(x_i, S_i)}_{i=1}^n$, where each $(x_i, S_i)$ is independently sampled from $p(x,S)$. 

\subsection{Deep Naive Loss for PLL}
The work of Jin and Gharhramani \cite{jin2003learning}, which is the first pioneering work on PLL, proposed a simple baseline method for PLL denoted as the `Naive model'. It is defined as follows:
\begin{align}
    \hat{\theta}=\argmax_{\theta\in\Theta}\sum_{i=1}^n\frac{1}{|S_i|}\sum_{y\in S_i}\log p \left(y|x_i; \theta \right).
    \label{eq:naivemodel}
\end{align}
We denote the \textit{naive loss} as the negative of the objective in the above.
In \cite{jin2003learning}, the authors proposed the disambiguation strategy as a better alternative to the naive model.
Moreover, many works on PLL \cite{zhang2015solving,tang2017confidence,feng2019partiala,DBLP:conf/aaai/YanG20} considered this naive model to be low-performing and it is still commonly believed that label disambiguation processes are crucial in achieving high-performance.

In this work, we propose the following differentiable loss to instantiate the naive loss with deep neural networks:
\begin{align}
    \label{eq:deepnaive}
    \hat{l}_n(\theta)&=-\frac{1}{n}\sum_{i=1}^n\log\rbra{\langle\overbar{S}_{\theta,i}\, S_i \rangle}, \\
    \label{eq:softmax}
    \overbar{S}_{\theta,i}&=\textsc{softmax}\rbra{\vec{f}_\theta(x_i)},
\end{align}
where $\vec{f}_\theta(x_i)\in \mathbb{R}^K$ is the output of the neural network. The softmax layer is used to make the outputs of the neural network lie in the probability simplex. One can see that the above loss is almost identical to the naive loss in (\ref{eq:naivemodel}) up to constant factors, hence we denote (\ref{eq:deepnaive}) as the \textit{deep naive loss} while a model trained from it is denoted as a \textit{deep naive model}. 

The above loss can be identified as a surrogate of the \textit{partial label risk} defined as follows:
\begin{align}
    R_p(\theta)=\expct_{(x,S)\sim p(x,S)}\I\rbra{h_\theta(x)\notin S},
    \label{eq:pllrisk}
\end{align}
where $\I\rbra{\cdot}$ is the indicator function. We denote $\hat{R}_{p,n}(\theta)$ as an empirical estimator of $R_p(\theta)$ over $n$ samples. When $h_\theta(x)=\argmax_{i} f_{\theta, i}(x)$, one can easily see that the deep naive loss (\ref{eq:deepnaive}) is a surrogate of the partial-label risk (\ref{eq:pllrisk}).

\subsection{Existing Theories of Generalization in PLL}
In this sub-section, we review two existing learning theories and their implications which may explain the effectiveness of deep naive models.
\subsubsection{EPRM Learnability}
Under a mild assumption on data distributions, Liu and Dietterich \cite{liu2014learnability} proved that minimizing an empirical partial label risk gives a correct classifier. 

Formally, they proved a finite sample complexity bound for the empirical partial risk minimizer (EPRM):
\begin{align}
    \hat{\theta}_n=\argmin_{\theta\in\Theta}\hat{R}_{p,n}(\theta),
\end{align}
under a mild distributional assumption called \textit{small ambiguity degree} condition. The \textit{ambiguity degree} \cite{cour2011learning} quantifies the hardness of a PLL problem and is defined as
\begin{align}
    \gamma=\sup_{\substack{(x,y)\in\X\times\Y,\\ \;\bar{y}\in\Y:p(x,y)>0,\;\bar{y}\neq y}} \Pr_{S\sim p(S|x,y)}\sbra{\bar{y}\in S}.
\end{align}
When $\gamma$ is less than 1, we say the \textit{small ambiguity degree condition} is satisfied. Intuitively, it measures how a specific non-ground-truth label co-occurs with a specific ground-truth label. When such \textit{distractor} labels co-occurs with a ground-truth label in every instance, it is impossible to disambiguate the label hence PLL is not EPRM learnable. With the mild assumption that $\gamma<1$, Liu and Ditterich showed the following sample complexity bound for PLL,
\begin{theorem}\normalfont{(PLL Sample complexity bound \cite{liu2014learnability})}.
    Suppose the ambiguity degree of a PLL problem is small, $0\leq\gamma<1$. Let $\eta=\log\frac{2}{1+\gamma}$ and $d_{\H}$ be the Natarajan dimension of the hypothesis space $\H$. Define
    \begin{align*}
        &n_0(\H,\eps,\del)=\\
        \frac{4}{\eta\eps}&\rbra{d_{\H}\rbra{\log{4d_\H}+2\log K+\log\frac{1}{\eta\eps}}+\log\frac{1}{\del}+1},
    \end{align*}
    then when $n>n_0(\H,\eps,\del)$, $\text{Err}(\hat{\theta}_n)<\eps$ with probability at least $1-\del$.
\label{thm:erm}
\end{theorem}
We denote this result as \textit{Empirical Partial Risk Minimization (EPRM) learnability}.

\subsubsection{Classifier-consistency}
A very recent work by Feng \textit{et al.} \cite{feng2020provably} proposed new PLL risk estimators by viewing the partial label generation process as a multiple complementary label generation process \cite{feng2020learning,cao2020multi}.
One of the proposed estimators is called \textit{classifier-consistent} (CC) risk $R_{cc}(\theta)$. For any multi-class loss function $\mathcal{L}: \mathbb{R}^K \times \Y \to \mathbb{R}_{+}$, $R_{cc}(\theta)$ it is defined as follows:
\begin{align}
R_{cc}(\theta) = \expct_{(x,S)\sim p(x,S)} \left[ \mathcal{L} \left( \mathbf{Q}^{\top} p \left(y | x ; \theta \right), s \right) \right],
\label{eq:cc}
\end{align}
\begin{table*}[]
\centering
\small
\def\arraystretch{1.0}
\begin{tabular}{c|cccc|c|cc}
Method & Lost       & MSRCv2     &  Soccer Player & Yahoo! News & Avg. Rank & Reference & Presented at   \\ \hline \hline
DNPL   & 81.1\rpm3.7\% (2) $\;\;$ & \textbf{54.4\rpm4.3\%} (1) $\;\;$ & 57.3\rpm1.4\% (2) $\;\;$ & \textbf{69.1\rpm0.9\%} (1) $\;\;$ & \textbf{1.50} & \multicolumn{2}{c}{This Work}  \\ \hline
CLPL   & 74.2\rpm3.8\% (7) $\bullet$ & 41.3\rpm4.1\% (12) $\bullet$ & 36.8\rpm1.0\% (12) $\bullet$ & 46.2\rpm0.9\% (12) $\bullet$ & 10.75 & \cite{cour2011learning} & JMLR 11 \\ \hline
CORD   & 80.6\rpm2.6\% (4) $\;\;$ & 47.4\rpm4.0\% (9) $\bullet$ & 45.7\rpm1.3\% (11) $\bullet$ & 62.4\rpm1.0\% (9) $\bullet$ & 8.25 & \cite{tang2017confidence} & AAAI 17 \\ \hline
ECOC   & 70.3\rpm5.2\% (9) $\bullet$ & 50.5\rpm2.7\% (6) $\bullet$ & 53.7\rpm2.0\% (7) $\bullet$ & 66.2\rpm1.0\% (5) $\bullet$ & 6.75 & \cite{zhang2017disambiguation} & TKDE 17 \\ \hline
GM-PLL & 73.7\rpm4.3\% (8) $\bullet$ & 53.0\rpm1.9\% (3) $\;\;$ & 54.9\rpm0.9\% (4) $\bullet$ & 62.9\rpm0.7\% (8) $\bullet$ & 5.75 & \cite{lyu2019gm} & TKDE 19 \\ \hline
IPAL   & 67.8\rpm5.3\% (10) $\bullet$ & 52.9\rpm3.9\% (4) $\;\;$ & 54.1\rpm1.6\% (5) $\bullet$ & 60.9\rpm1.1\% (10) $\bullet$ & 7.25 & \cite{zhang2015solving} & AAAI 15 \\ \hline
PL-BLC & 80.6\rpm3.2\% (4) $\;\;$ & 53.6\rpm3.7\% (2) $\;\;$ & 54.0\rpm0.8\% (6) $\bullet$ & 67.9\rpm0.5\% (2) $\bullet$ & 3.50 & \cite{DBLP:conf/aaai/YanG20} & AAAI 20  \\ \hline
PL-LE  & 62.9\rpm5.6\% (11) $\bullet$ & 49.9\rpm3.7\% (7) $\bullet$ & 53.6\rpm2.0\% (8) $\bullet$ & 65.3\rpm0.6\% (6) $\bullet$ & 8.00 & \cite{xu2019partial} & AAAI 19 \\ \hline
PLKNN  & 43.2\rpm5.1\% (12) $\bullet$ & 41.7\rpm3.4\% (11) $\bullet$ & 49.5\rpm1.8\% (10) $\bullet$ & 48.3\rpm1.1\% (11) $\bullet$ & 11.00 & \cite{hullermeier2006learning} & IDA 06 \\ \hline
PRODEN & \textbf{81.6\rpm3.5\%} (1) $\;\;$ & 43.4\rpm3.3\% (10) $\bullet$ & 55.3\rpm5.6\% (3) $\bullet$ & 67.5\rpm0.7\% (3) $\bullet$ & 4.25 & \cite{lv2020progressive} & ICML 20 \\ \hline
SDIM   & 80.1\rpm3.1\% (5) $\;\;$ & 52.0\rpm3.7\% (5) $\;\;$ & \textbf{57.7\rpm1.6\%} (1) $\;\;$ & 66.3\rpm1.3\% (4) $\bullet$ & 3.75 & \cite{feng2019partiala} & IJCAI 19 \\ \hline
SURE   & 78.0\rpm3.6\% (6) $\bullet$ & 48.1\rpm3.6\% (8) $\bullet$ & 53.3\rpm1.7\% (9) $\bullet$ & 64.4\rpm1.5\% (7) $\bullet$ & 7.50 & \cite{feng2019partialb} & AAAI 19
\end{tabular}
\caption{Benchmark results (mean accuracy$\rpm$std) on the real-world datasets. Numbers in parenthesis represent rankings of comparing methods and the sixth column is the average rankings. Best methods are emphasized in boldface. $\bullet/\circ$ indicates whether our method (DNPL) is better/worse than the comparing methods with respect to unpaired Welch $t$-test at $5\%$ significance level.}
\label{table:real-bench}
\end{table*}

where $\mathbf{Q} \in \mathbb{R}^{K \times K}$ is a label transition matrix in the context of multiple complementary label learning, $s$ is a uniformly randomly chosen label from $S$.
$\hat{R}_{cc,n}(\theta)$ is denoted as empirical risk of Eq. \ref{eq:cc}.

Feng \textit{et al.}'s main contribution is to prove an estimation error bound for the CC risk (\ref{eq:cc}).
Let $\hat{\theta}_n = \argmin_{\theta \in \Theta} \hat{R}_{cc,n}(\theta)$ and $\theta^{\star} = \argmin_{\theta \in \Theta} R_{cc}(\theta)$ denote the empirical and the true minimizer, respectively.
Additionally, $\H_y$ refers the model hypothesis space for label $y$.
Then, the estimation error bound for the CC risk is given as
\begin{theorem}\normalfont{(Estimation error bound for the CC risk \cite{feng2020provably})}.
\label{cls_bound}
Assume the loss function $\mathcal{L} \left( \mathbf{Q}^{\top} p \left(y | x ; \theta \right), s \right)$ is $\rho$-Lipschitz with respect to the first augment in the 2-norm and upper-bounded by $M$. Then, for any $\delta>0$, with probability at least $1-\delta$,
\begin{gather}
\nonumber
\textstyle R_{cc}(\hat{\theta}_n) - R_{cc}(\theta^\star)\leq
8\rho\sum\nolimits_{y=1}^k{\mathfrak{R}}_n(\mathcal{H}_y)+2M\sqrt{\frac{\log\frac{2}{\delta}}{2n}},
\end{gather}
where $\mathfrak{R}_n (\mathcal{H}_y)$ refers the expected Rademacher complexity of the hypothesis space for the label $y$, $\mathcal{H}_y$, with sample size $n$.
\label{thm:cc}
\end{theorem}
If the uniform label transition probability is assumed i.e., $\mathbf{Q}_{ij} = \delta_{ij} \I\rbra{j\in S_j}/\left({2^{K-1} - 1} \right)$, Eq. \ref{eq:cc} becomes equivalent to our deep naive loss (Eq. \ref{eq:deepnaive}) up to some constant factors.
Hence, Theorem \ref{thm:erm} and \ref{thm:cc} give generalization bounds on the partial risk and the CC risk (same as Eq. \ref{eq:deepnaive}) respectively.

\subsection{Alternative Explanation of Generalization in DNPL}
Since the work of \cite{zhang2017understanding}, the mystery of deep learning's generalization ability has been widely investigated in the standard supervised learning setting. While it is still not fully understood why over-parametrized deep neural networks generalize well, several studies are suggesting that deep learning models are inherently biased toward simple functions \cite{valle2018deep, de2019random}. 
Especially, Valle-Perez et al. \cite{valle2018deep} empirically observed that solutions from stochastic gradient descent (SGD) are biased toward neural networks with smaller complexity. They observed the following universal scaling behavior in the output distribution $p(\theta)$ of SGD:
\begin{align}
    p(\theta)\lesssim e^{-a C(\theta) + b},
\label{eq:outp}
\end{align}
where $C(\theta)$ is a computable proxy of (uncomputable) Kolmogorov complexity and $a$, $b$ are $\theta$-independent constants.
One example of complexity measure $C(\theta)$ is Lempel-Ziv complexity \cite{valle2018deep} which is roughly the length of compressed $\theta$ with ZIP compressor.

In the deep naive PLL, the model parameter is a minimizer of the empirical partial label risk $\hat{R}_{p,n}(\theta)$ (Eq. \ref{eq:pllrisk}). The minima of $\hat{R}_{p,n}(\theta)$ is wide because there are many model parameters perfectly fit to given partially labeled examples. The support of SGD's output distribution will lie in this wide minima. According to Eq. \ref{eq:outp}, this distribution is heavily biased toward parameters with small complexities. One crucial observation is that models fitting inconsistent labels will generally have large complexities since they have to memorize each example. According to Eq. \ref{eq:outp}, such models are exponentially unlikely to be outputted by SGD. Hence the most likely output of the deep naive PLL method is a classifier with small error.
As a result, the implications of both Theorem \ref{thm:erm} and \ref{thm:cc} appear to be empirically correct in spite of their vacuity of model complexity.

\section{Experiments}
\label{sec:pagestyle}
\begin{figure*}[h]
\begin{multicols}{2}
    \includegraphics[width=0.92\linewidth]{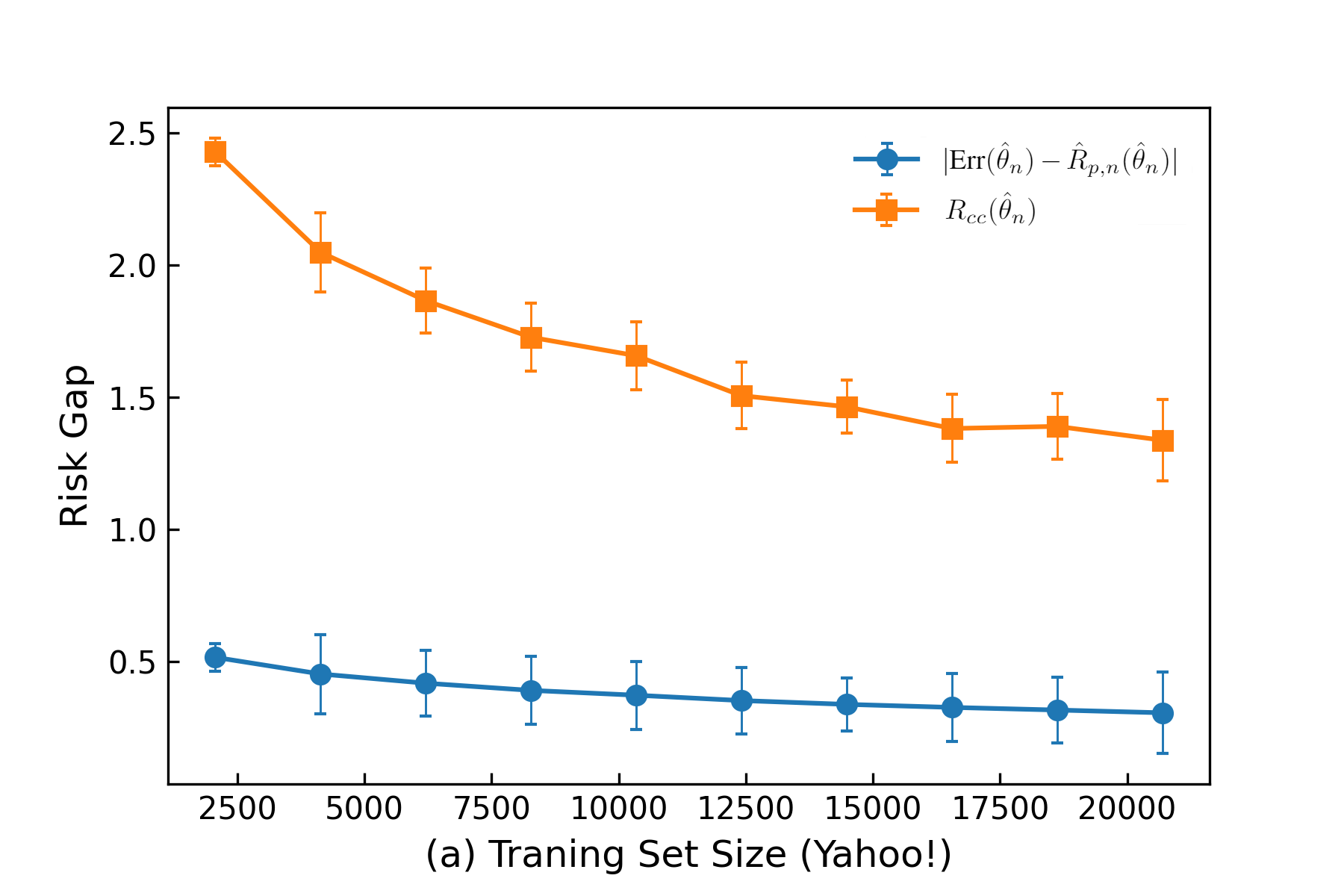}\par
    \includegraphics[width=0.92\linewidth]{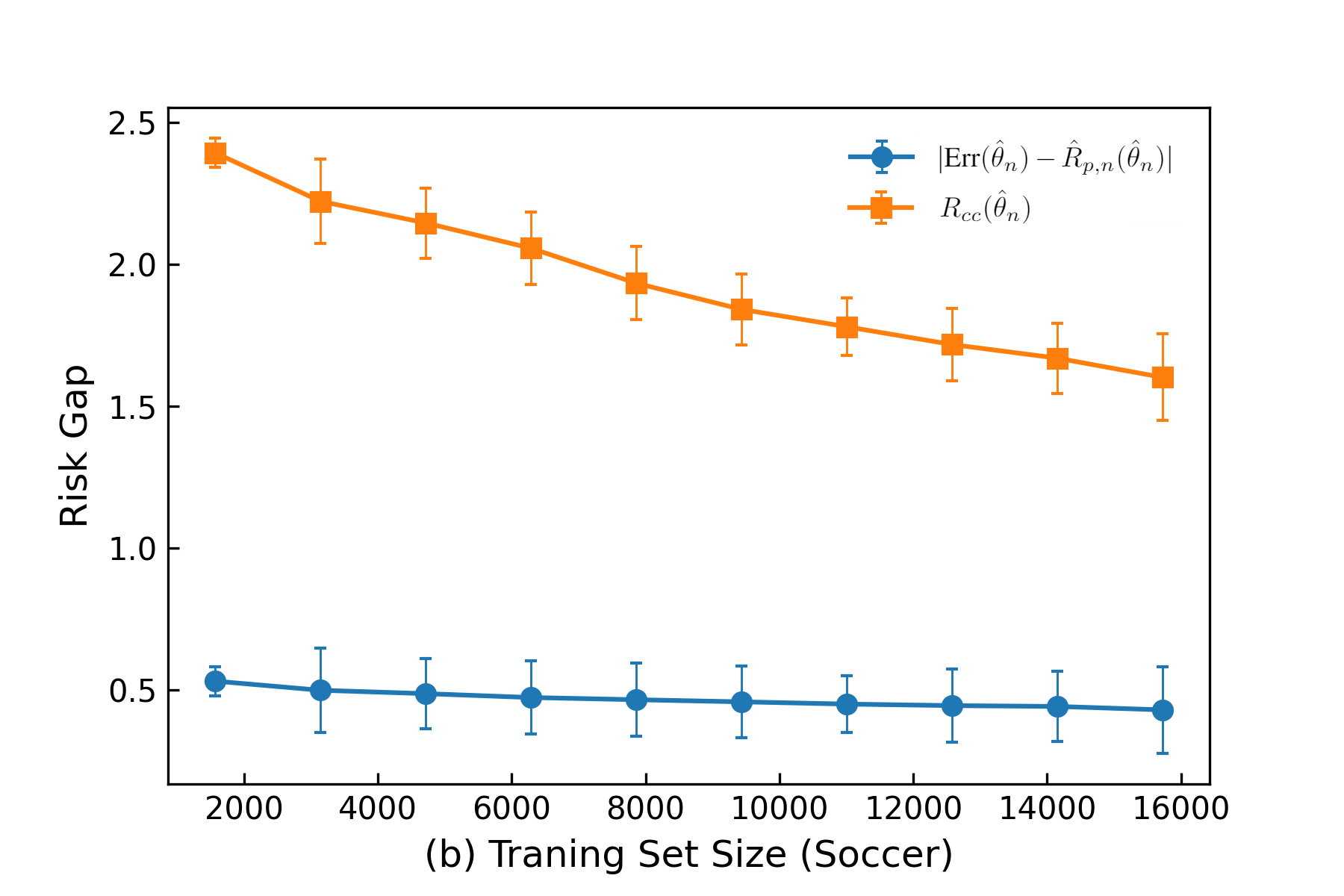}\par
\end{multicols}
\vspace*{-8mm}
\caption{Generalization gaps with respect to training set size for (a) Yahoo! dataset and (b) Soccer dataset are shown. Error bars represent STDs over 10 repeated experiments. Note that we went through the same experiment process for the other two smaller datasets (Lost / MSRCv2), but these results were omitted because of the same tendency.}
\label{fig:gengap}
\end{figure*}
In this section we give the readers two points.
First, deep neural network classifiers trained with the naive loss can achieve competitive performance in real-world benchmarks.
Second, the generalization gaps of trained classifiers effectively decrease with respect to the increasing training set size.
\subsection{Benchmarks on Real-world PLL Datasets}
\subsubsection{Datasets and Comparing Methods}
We use four real-world datasets including \textit{Lost} \cite{panis2016overview}, \textit{MSRCv2} \cite{liu2012conditional}, \textit{Soccer Player} \cite{zeng2013learning}, and \textit{Yahoo! News} \cite{guillaumin2010multiple}.
All real-world datasets can be found in this website\footnote{ http://palm.seu.edu.cn/zhangml/}.
We denote the suggested method as Deep Naive Partial label Learning (DNPL). We compare DNPL with eleven baseline methods.
There are eight parametric methods: CLPL, CORD \cite{tang2017confidence}, ECOC, PL-BLC \cite{DBLP:conf/aaai/YanG20}, PL-LE \cite{xu2019partial}, PRODEN, SDIM \cite{feng2019partiala}, SURE, and three non-parametric methods: GM-PLL \cite{lyu2019gm}, IPAL, PLKNN.
Note that both CORD and PL-BLC are deep learning-based PLL methods which includes label identification or mean-teaching techniques.

\subsubsection{Models and Hyperparameters}
We employ a neural network of the following architecture: $d_\text{in}-512-256-d_\text{out}$, where numbers represent dimensions of layers and $d_\text{in}$ ($d_\text{out}$) is input (output) dimension. The neural network have the same size as that  of PL-BLC.
Batch normalization \cite{pmlr-v37-ioffe15} is applied after each layer followed by ELU activation layer \cite{clevert2016fast}.
Yogi optimizer \cite{zaheer2018adaptive} is used with fixed learning rate $10^{-3}$ and default momentum parameters $\rbra{0.9,\,0.999}$.

\subsubsection{Benchmark Results}
Table \ref{table:real-bench} reports means and standard deviations of observed accuracies. Accuracies of the naive model are measured over 5 repeated 10-fold cross-validation and accuracies of others are measured over 10-fold cross-validation.

The benchmark results indicate that DNPL achieves state-of-the-art performances over all four datasets. Especially, DNPL outperforms PL-BLC which uses a neural network of the same size as ours on those datasets. Unlike PL-BLC or CORD, DNPL does not need computationally expensive processes like label identification and mean-teaching.
This means that by simply borrowing our surrogate loss to the deep learning classifier, we can build a sufficiently competitive PLL model.

Observing that for \textit{Soccer Player} and \textit{Yahoo! News} datasets, DNPL outperforms almost all of the comparing methods. Regarding the large-scale and high-dimensional nature of \textit{Soccer Player} and \textit{Yahoo! News} datasets comparing to other datasets, this observation suggests that DNPL has its advantage on large-scale, high-dimensional datasets.

\subsection{Generalization Gaps of Deep Naive PLL}
In this section, we empirically show that conventional learning theories (Theorem \ref{thm:erm}, \ref{thm:cc}) cannot explain the learning behaviors of DNPL.
Figure \ref{fig:gengap} shows how the gap $|\text{Err}(\hat{\theta}_n) - \hat{R}_{p,n}(\hat{\theta}_n)|$ and the CC risk\footnote{We have always observed that with our over-parameterized neural network zero risk can be achieved for $R_{cc}(\theta^\star)$. Therefore, we omit this term.} $R_{cc}(\hat{\theta}_n)$ decreases as dataset size $n$ increases.\ We observe that gap closing behaviors despite the neural networks are over-parametrized, i.e., \# of parameters $\sim 10^5 >>$ the training set size $\sim 10^4$.

\section{Conclusions}
This work showed that a simple naive loss is applicable in training high-performance deep classifiers with partially labeled examples. Moreover, this method does not require any label disambiguation or explicit regularization. Our observations indicate that the deep naive method's unreasonable effectiveness cannot be explained by existing learning theories. These raise interesting questions deserving further studies: 1) To what extent does the label disambiguation help learning with partial labels? 2) How deep learning generalizes in partial label learning?

\vfill\clearpage 

\bibliographystyle{IEEEbib}
\bibliography{refs}

\begin{thebibliography}{10}

\bibitem{hsu2019multi}
Yen-Chang Hsu, Zhaoyang Lv, Joel Schlosser, Phillip Odom, and Zsolt Kira,
\newblock ``Multi-class classification without multi-class labels,''
\newblock in {\em ICLR}, 2019.

\bibitem{kiryo2017positive}
Ryuichi Kiryo, Gang Niu, Marthinus~C Du~Plessis, and Masashi Sugiyama,
\newblock ``Positive-unlabeled learning with non-negative risk estimator,''
\newblock in {\em NeurIPS}, 2017.

\bibitem{kaji2018multi}
Hirotaka Kaji, Hayato Yamaguchi, and Masashi Sugiyama,
\newblock ``Multi task learning with positive and unlabeled data and its
  application to mental state prediction,''
\newblock in {\em ICASSP}, 2018.

\bibitem{maron1998framework}
Oded Maron and Tom{\'a}s Lozano-P{\'e}rez,
\newblock ``A framework for multiple-instance learning,''
\newblock in {\em NeurIPS}, 1998.

\bibitem{wang2019comparison}
Yun Wang, Juncheng Li, and Florian Metze,
\newblock ``A comparison of five multiple instance learning pooling functions
  for sound event detection with weak labeling,''
\newblock in {\em ICASSP}, 2019.

\bibitem{jin2003learning}
Rong Jin and Zoubin Ghahramani,
\newblock ``Learning with multiple labels,''
\newblock in {\em NeurIPS}, 2003.

\bibitem{luo2010learning}
Jie Luo and Francesco Orabona,
\newblock ``Learning from candidate labeling sets,''
\newblock in {\em NeurIPS}, 2010.

\bibitem{liu2012conditional}
Liping Liu and Thomas~G Dietterich,
\newblock ``A conditional multinomial mixture model for superset label
  learning,''
\newblock in {\em NeurIPS}, 2012.

\bibitem{zeng2013learning}
Zinan Zeng, Shijie Xiao, Kui Jia, Tsung-Han Chan, Shenghua Gao, Dong Xu, and
  Yi~Ma,
\newblock ``Learning by associating ambiguously labeled images,''
\newblock in {\em CVPR}, 2013.

\bibitem{hullermeier2006learning}
Eyke H{\"u}llermeier and J{\"u}rgen Beringer,
\newblock ``Learning from ambiguously labeled examples,''
\newblock {\em Intell Data Anal}, 2006.

\bibitem{cour2011learning}
Timothee Cour, Ben Sapp, and Ben Taskar,
\newblock ``Learning from partial labels,''
\newblock {\em JMLR}, 2011.

\bibitem{zhang2015solving}
Min-Ling Zhang and Fei Yu,
\newblock ``Solving the partial label learning problem: An instance-based
  approach,''
\newblock in {\em AAAI}, 2015.

\bibitem{tang2017confidence}
Cai-Zhi Tang and Min-Ling Zhang,
\newblock ``Confidence-rated discriminative partial label learning,''
\newblock in {\em AAAI}, 2017.

\bibitem{feng2019partiala}
Lei Feng and Bo~An,
\newblock ``Partial label learning by semantic difference maximization,''
\newblock in {\em IJCAI}, 2019.

\bibitem{DBLP:conf/aaai/YanG20}
Yan Yan and Yuhong Guo,
\newblock ``Partial label learning with batch label correction,''
\newblock in {\em AAAI}, 2020.

\bibitem{valle2018deep}
Guillermo Valle-Perez, Chico~Q Camargo, and Ard~A Louis,
\newblock ``Deep learning generalizes because the parameter-function map is
  biased towards simple functions,''
\newblock in {\em ICLR}, 2018.

\bibitem{liu2014learnability}
Liping Liu and Thomas Dietterich,
\newblock ``Learnability of the superset label learning problem,''
\newblock in {\em ICML}, 2014.

\bibitem{feng2020provably}
Lei Feng, Jiaqi Lv, Bo~Han, Miao Xu, Gang Niu, Xin Geng, Bo~An, and Masashi
  Sugiyama,
\newblock ``Provably consistent partial-label learning,''
\newblock in {\em NeurIPS}, 2020.

\bibitem{feng2020learning}
Lei Feng and Bo~An,
\newblock ``Learning from multiple complementary labels,''
\newblock in {\em ICML}, 2020.

\bibitem{cao2020multi}
Yuzhou Cao and Yitian Xu,
\newblock ``Multi-complementary and unlabeled learning for arbitrary losses and
  models,''
\newblock in {\em ICML}, 2020.

\bibitem{zhang2017disambiguation}
Min-Ling Zhang, Fei Yu, and Cai-Zhi Tang,
\newblock ``Disambiguation-free partial label learning,''
\newblock {\em IEEE Trans Knowl Data Eng}, 2017.

\bibitem{lyu2019gm}
Gengyu Lyu, Songhe Feng, Tao Wang, Congyan Lang, and Yidong Li,
\newblock ``Gm-pll: Graph matching based partial label learning,''
\newblock {\em IEEE Trans Knowl Data Eng}, 2019.

\bibitem{xu2019partial}
Ning Xu, Jiaqi Lv, and Xin Geng,
\newblock ``Partial label learning via label enhancement,''
\newblock in {\em AAAI}, 2019.

\bibitem{lv2020progressive}
Jiaqi Lv, Miao Xu, Lei Feng, Gang Niu, Xin Geng, and Masashi Sugiyama,
\newblock ``Progressive identification of true labels for partial-label
  learning,''
\newblock in {\em ICML}, 2020.

\bibitem{feng2019partialb}
Lei Feng and Bo~An,
\newblock ``Partial label learning with self-guided retraining,''
\newblock in {\em AAAI}, 2019.

\bibitem{zhang2017understanding}
Chiyuan Zhang, Samy Bengio, Moritz Hardt, Benjamin Recht, and Oriol Vinyals,
\newblock ``Understanding deep learning requires rethinking generalization,''
\newblock in {\em ICLR}, 2017.

\bibitem{de2019random}
Giacomo De~Palma, Bobak Kiani, and Seth Lloyd,
\newblock ``Random deep neural networks are biased towards simple functions,''
\newblock in {\em NeurIPS}, 2019.

\bibitem{panis2016overview}
Gabriel Panis, Andreas Lanitis, Nicholas Tsapatsoulis, and Timothy~F Cootes,
\newblock ``Overview of research on facial ageing using the \text{FG-NET}
  ageing database,''
\newblock {\em IET Biometrics}, 2016.

\bibitem{guillaumin2010multiple}
Matthieu Guillaumin, Jakob Verbeek, and Cordelia Schmid,
\newblock ``Multiple instance metric learning from automatically labeled bags
  of faces,''
\newblock in {\em ECCV}, 2010.

\bibitem{pmlr-v37-ioffe15}
Sergey Ioffe and Christian Szegedy,
\newblock ``Batch normalization: Accelerating deep network training by reducing
  internal covariate shift,''
\newblock in {\em ICML}, 2015.

\bibitem{clevert2016fast}
Djork-Arn{\'e} Clevert, Thomas Unterthiner, and Sepp Hochreiter,
\newblock ``Fast and accurate deep network learning by exponential linear units
  (\text{ELU}s),''
\newblock in {\em ICLR}, 2016.

\bibitem{zaheer2018adaptive}
Manzil Zaheer, Sashank Reddi, Devendra Sachan, Satyen Kale, and Sanjiv Kumar,
\newblock ``Adaptive methods for nonconvex optimization,''
\newblock in {\em NeurIPS}, 2018.

\end{thebibliography}

\end{document}